\documentclass[11pt,a4paper]{article}
\usepackage[hyperref]{conll-2019}
\usepackage{times}
\usepackage{latexsym}
\usepackage{graphicx}
\usepackage{booktabs} 

\usepackage{url}

\aclfinalcopy 

\title{Evaluating the Cross-Lingual Effectiveness of \\ Massively Multilingual Neural Machine Translation}

\author{Aditya Siddhant, Melvin Johnson, Henry Tsai, Naveen Arivazhagan,\\
\textbf{Jason Riesa, Ankur Bapna, Orhan Firat, Karthik Raman} \\
Google Research \\
\{adisid, melvinp, henrytsai, navari, reisa, ankurbpn, orhanf, karthikraman\}@google.com}

\date{}

\begin{document}
\maketitle
\begin{abstract}
%
  
  
  
    The recently proposed massively multilingual neural machine translation (NMT) system has been shown to be capable of translating over 100 languages to and from English within a single model \cite{aharoni2019massively}. Its improved translation performance on low resource languages hints at potential cross-lingual transfer capability for downstream tasks. In this paper, we evaluate the cross-lingual effectiveness of representations from the encoder of a massively multilingual NMT model on 5 downstream classification and sequence labeling tasks covering a diverse set of over 50 languages. We compare against a strong baseline, multilingual BERT (mBERT) \cite{devlin2018bert}, in different cross-lingual transfer learning scenarios and show gains in zero-shot transfer in 4 out of these 5 tasks.

\end{abstract}





\section{Introduction}
English has an abundance of labeled data that can be used for various Natural Language Processing (NLP) tasks, such as part-of-speech tagging (POS), named entity recognition (NER), and natural language inference (NLI). This richness of labeled data manifests itself as a boost in accuracy in the current era of data-hungry deep learning algorithms. 
However, the same is not true for many other languages where task specific data is scarce and expensive to acquire. This motivates the need for cross-lingual transfer learning -- the ability to leverage the knowledge from task specific data available in one or more languages to solve that task in languages with little or no task-specific data. 

Recent progress in NMT has enabled one to train multilingual systems that support translation from multiple source languages into multiple target languages within a single model~\cite{firat2016multi,johnson2017google,aharoni2019massively}. Such multilingual NMT (mNMT) systems often demonstrate large improvements in translation quality on low resource languages. This positive transfer originates from the model's ability to learn  representations which are transferable across languages. Previous work has shown that these representations can then be used for cross-lingual transfer in other downstream NLP tasks - albeit on only a pair of language pairs~\cite{eriguchi2018zero}, or by limiting the decoder to use a pooled vector representation of the entire sentence from the encoder \cite{artetxe2018massively}.

In this paper we scale up the number of translation directions used in the NMT model to include 102 languages to and from English. Unlike \citet{artetxe2018massively}, we do not apply any restricting operations such as pooling while training mNMT which allows us to obtain token level representations making it possible to transfer them to sequence tagging tasks as well. We find that mNMT models trained using plain translation losses can out of the box emerge as competitive alternatives to other methods at the forefront of cross-lingual transfer learning \cite{devlin2018bert,artetxe2018massively}

Our contributions in this paper are threefold:
\begin{itemize}
\item We use representations from a Massively Multilingual Translation Encoder (MMTE) that can handle 103 languages to achieve cross-lingual transfer on 5 classification and sequence tagging tasks spanning more than 50 languages.
\item We compare MMTE to mBERT in different cross-lingual transfer scenarios including zero-shot, few-shot, fine-tuning, and feature extraction scenarios.
\item We outperform the state-of-the-art on zero-shot cross-lingual POS tagging [Universal Dependencies 2.3 dataset \cite{nivre2018universal}], intent classification \cite{schuster2018cross}, and achieve results comparable to state-of-the-art on document classification [ML-Doc dataset \cite{SCHWENK18.658}].
\end{itemize}


The remainder of this paper is organized as follows. Section \ref{sec: mmte} describes our MMTE model in detail and points out its differences from mBERT. All experimental details, results and analysis are given in Sections \ref{sec: results} and \ref{sec: analysis}. This is followed by a discussion of related work. In Section~\ref{conclusion}, we summarize our findings and present directions for future research. We emphasize that the primary motivation of the paper is not to challenge the state-of-the-art but instead to investigate the effectiveness of representations learned from an mNMT model in various transfer-learning settings.

\section{Massively Multilingual Neural Machine Translation Model} \label{sec: mmte}

In this section, we describe our massively multilingual NMT system. Similar to BERT, our transfer learning setup has two distinct steps: \textit{pre-training} and \textit{fine-tuning}. During pre-training, the NMT model is trained on large amounts of parallel data to perform translation.  During fine-tuning, we initialize our downstream model with the pre-trained parameters from the encoder of the NMT system, and then all of the parameters are fine-tuned using labeled data from the downstream tasks.

\subsection{Model Architecture}
We train our Massively Multilingual NMT system using the Transformer architecture \citep{vaswani2017attention} in the open-source implementation under the Lingvo framework \citep{lingvo}. We use a larger version of Transformer Big containing 375M parameters (6 layers, 16 heads, 8192 hidden dimension) \citep{chen-EtAl:2018:Long1}, and a shared source-target sentence-piece model (SPM)\footnote{https://github.com/google/sentencepiece} \cite{kudo2018sentencepiece} vocabulary with 64k individual tokens. All our models are trained with Adafactor \citep{shazeer2018adafactor} with momentum factorization, a learning rate schedule of (3.0, 40k)\footnote{The shorthand form (3.0, 40k) corresponds to a learning rate of 3.0, with 40k warm-up steps for the schedule, which is decayed with the inverse square root of the number of training steps after warm-up.} and a per-parameter norm clipping threshold of 1.0. The encoder of this NMT model comprises approximately 190M parameters and is subsequently used for fine-tuning.

\subsection{Pre-training}

\paragraph{Objective}
We train a massively multilingual NMT system which is capable of translating between a large number of language pairs at the same time by optimizing the translation objective between language pairs. To train such a multilingual system within a single model, we use the strategy proposed in \cite{johnson2017google} which suggests prepending a target language token to every source sequence to be translated. This simple and effective strategy enables us to share the encoder, decoder, and attention mechanisms across all language pairs.

\begin{figure}[t!]
\begin{center}
\includegraphics[scale=0.35]{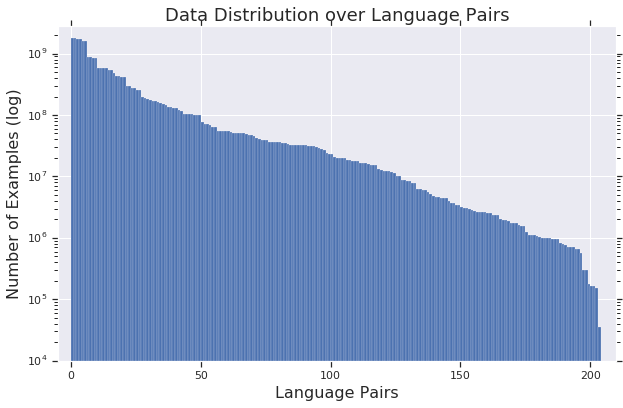}
\caption{Per language pair data distribution of the dataset used for our multilingual NMT model. The y-axis depicts the number of training examples available per language pair on a logarithmic scale. Dataset sizes range from 35k for the lowest resource language pairs to 2 billion for the largest.}
\label{fig:data}
\end{center}
\end{figure}

\paragraph{Data}
We train our multilingual NMT system on a massive scale, using an in-house corpus generated by crawling and extracting parallel sentences from the web \cite{uszkoreit2010large}. This corpus contains parallel documents for 102 languages, to and from English, comprising a total of 25 billion sentence pairs. The number of parallel sentences per language in our corpus ranges from around 35 thousand to almost 2 billion. Figure \ref{fig:data} illustrates the data distribution for all 204 language pairs used to train the NMT model. Language ids for all the languages are also provided in supplementary material.

\paragraph{Data sampling policy}
Given the wide distribution of data across language pairs, we used a temperature based data balancing strategy. For a given language pair, $l$, let $D_l$ be the size of the available parallel corpus. Then if we adopt a naive strategy and sample from the union of the datasets, the probability of the sample being from language pair $l$ will be $p_l=\frac{D_l}{\Sigma_lD_l}$. However, this strategy would starve low resource language pairs. To control for the ratio of samples from different language pairs, we sample a fixed number of sentences from the training data, with the probability of a sentence belonging to language pair $l$ being proportional to $p_l^{\frac{1}{T}}$, where $T$ is the sampling temperature. As a result, $T=1$ would correspond to a true data distribution, and, $T=100$ yields an (almost) equal number of samples for each language pair (close to a uniform distribution with over-sampling for low-resource language-pairs). We set $T=5$ for a balanced sampling strategy. To control the contribution of each language pair when constructing the vocabulary, we use the same temperature based sampling strategy with $T=5$. Our SPM vocabulary has a character coverage of $0.999995$.

\paragraph{Model quality}
We use BLEU score~\cite{papineni2002bleu} to evaluate the quality of our translation model(s). Our mNMT model performs worse than the bilingual baseline on high resource language pairs but improves upon it on low resource language pairs. The average drop in BLEU score on 204 language pairs as compared to bilingual baselines is just 0.25 BLEU. This is impressive considering we are comparing one multilingual model to 204 different bilingual models. Table \ref{bleuscore table} compares the BLEU scores achieved by mNMT to that of the bilingual baselines on 10 representative language pairs.\footnote{We chose a diverse set of language pairs with varying language families, scripts, and dataset sizes.} These scores are obtained on an internal evaluation set which contains around 5k examples per language pair.

\begin{table}[]
\centering
\resizebox{\linewidth}{!}{
\begin{tabular}{c|cc|c|cc}
\toprule
\multicolumn{1}{l}{} & \multicolumn{1}{l}{baseline*} & \multicolumn{1}{l}{mNMT} & \multicolumn{1}{l}{} & \multicolumn{1}{l}{baseline*} & \multicolumn{1}{l}{mNMT} \\
\midrule
ur-en & \textbf{27.84} & 27.24          & en-bg & \textbf{31.31} & 29.36          \\
mr-en & 27.81          & \textbf{28.61} & en-es & \textbf{35.23} & 34.35          \\
be-en & 24.23          & \textbf{24.66} & en-sw & 18.73          & \textbf{19.79} \\
ru-en & 26.54          & \textbf{26.56} & en-pt & 37.19          & \textbf{37.41} \\
fr-en & \textbf{37.49} & 34.02          & en-hi & 16.46          & \textbf{16.63}  \\
\end{tabular}}
\caption{BLEU scores on ten language pairs with mNMT. *baseline refers to a transformer based bilingual NMT model trained on only one language pair.} \label{bleuscore table}
\end{table}

\subsection{Fine-tuning mNMT Encoder}
Fine-tuning involves taking the encoder of our mNMT model, named Massively Multilingual Translation Encoder (MMTE), and adapting it to the downstream task. For tasks which involve single input, the text is directly fed into the encoder. For tasks such as entailment which involve input pairs, we concatenate the two inputs using a separator token and pass this through the encoder. For each downstream task, the inputs and outputs are passed through the encoder and we fine-tune all the parameters end-to-end. The encoder encodes the input through the stack of Transformer layers and produces representations for each token at the output. For sequence tagging tasks, these token level representations are individually fed into a task-specific output layer. For classification or entailment tasks, we apply max-pooling on the token level representations and feed this into the task-specific output layer.

It should be noted that fine-tuning is relatively inexpensive and fast. All of the results can be obtained within a few thousand gradient steps. The individual task-specific modeling details are described in detail in section \ref{sec: results}.
It is also important to note that while the encoder, the attention mechanism, and the decoder of the model are trained in the pre-training phase, only the encoder is used during fine-tuning.

\subsection{Differences with mBERT}
We point out some of the major difference between mBERT and MMTE are:
\begin{itemize} 
\item mBERT uses two unsupervised pre-training objectives called masked language modeling (MLM) and next sentence prediction (NSP) which are both trained on monolingual data in 104 languages. MMTE on the other hand uses parallel data in 103 languages (102 languages to and from English) for supervised training with negative log-likelihood as the loss. It should be noted that mBERT uses clean Wikipedia data while MMTE is pre-trained on noisy parallel data from the web.
\item mBERT uses 12 transformer layers, 12 attention heads, 768 hidden dimensions and has 178M parameters while MMTE uses 6 transformer layers, 16 attention heads, and 8196 hidden dimensions with 190M parameters. Note that, the effective capacity of these two models cannot easily be compared by simply counting number of parameters, due to the added characteristic complexity with depth and width.
\item MMTE uses SPM to tokenize input with 64k vocabulary size while mBERT uses a Wordpiece model \citep{wu2016google} with 110k vocabulary size. 
\end{itemize}

\section{Experiments and Results} \label{sec: results}
As stated earlier, we use MMTE to perform downstream cross-lingual transfer on 5 NLP tasks. These include 3 classification tasks: NLI (XNLI dataset), document classification (MLDoc dataset) and intent classification, and 2 sequence tagging tasks: POS tagging and NER. We detail all of the experiments in this section.

\subsection{XNLI: Cross-lingual NLI}
XNLI is a popularly used corpus for evaluating cross-lingual sentence classification. It contains data in 15 languages \cite{conneau2018xnli}. Evaluation is based on classification accuracy for pairs of sentences as one of entailment, neutral, or contradiction. We feed the text pair separated by a special token into MMTE and add a small network on top of it to build a classifier. This small network consists of a pre-pool feed-forward layer with 64 units, a max-pool layer which pools word level representations to get the sentence representation, and a post-pool feed-forward layer with 64 units. The optimizer used is Adafactor with a learning rate schedule of (0.2, 90k). The classifier is trained on English only and evaluated on all the 15 languages. Results are reported in Table \ref{tab:xnli}. Please refer to Appedix Table 1 for language names associated with the codes.

\begin{table}[h]
\centering
\resizebox{.7\linewidth}{!}{
\begin{tabular}{c|cc|c}
\toprule
    & mBERT*         & MMTE          & SOTA$^\dagger$ \\
\midrule
en  & \textbf{82.1} & 79.6          & 85.0 \\
ar  & 64.9          & \textbf{64.9} & 73.1 \\
bg  & 68.9          & \textbf{70.4} & 77.4 \\
de  & \textbf{71.1} & 68.2          & 77.8 \\
el  & 66.4          & \textbf{67.3} & 76.6 \\
es  & \textbf{74.3} & 71.6          & 78.9 \\
fr  & \textbf{73.8} & 69.5          & 78.7 \\
hi  & 60.0          & \textbf{63.5} & 69.6 \\
ru  & \textbf{69.0} & 66.2          & 75.3 \\
sw  & 50.4          & \textbf{61.9} & 68.4 \\
th  & 55.8          & \textbf{66.2} & 73.2 \\
tr  & 61.6          & \textbf{63.6} & 72.5 \\
ur  & 58.0          & \textbf{60.0} & 67.3 \\
vi  & 69.5          & \textbf{69.7} & 76.1 \\
zh  & \textbf{69.3} & 69.2          & 76.5 \\
\midrule
avg & 66.3          & \textbf{67.5} & 75.1 \\
\bottomrule
\end{tabular}} 
\caption{Accuracies on the test set of the XNLI dataset. *mBERT numbers have been taken from \citet{wu2019beto}. $\dagger$ SOTA in the last column refers to the MLM + translation language modeling (TLM) results reported in \citet{lample2019cross}.} \label{tab:xnli}
\end{table}

MMTE outperforms mBERT on 9 out of 15 languages and by 1.2 points on average. BERT achieves excellent results on English, outperforming our system by 2.5 points but its zero-shot cross-lingual transfer performance is weaker than MMTE. We see most gains in low resource languages such as ar, hi, ur, and sw. MMTE however falls short of the current state-of-the-art (SOTA) on XNLI~\citep{lample2019cross}. We hypothesize this might be because of 2 reasons: (1) They use only the 15 languages associated with the XNLI task for pre-training their model, and (2) They use both monolingual and parallel data for pre-training while we just use parallel data. We confirm our first hypothesis later in Section~\ref{sec: analysis} where we see that decreasing the number of languages in mNMT improves the performance on XNLI.

\subsection{MLDoc: Document Classification}
MLDoc is a balanced subset of the Reuters corpus covering 8 languages for document classification \citep{SCHWENK18.658}. This is a 4-way classification task of identifying topics between CCAT (Corporate/Industrial), ECAT (Economics), GCAT (Government/Social), and MCAT (Markets). Performance is evaluated based on classification accuracy. We split the document using the sentence-piece model and feed the first 200 tokens into the encoder for classification. The task-specific network and the optimizer used is same as the one used for XNLI. Learning rate schedule is (0.2,5k). We perform both in-language and zero-shot evaluation. The in-language setting has training, development and test sets from the language. In the zero-shot setting, the train and dev sets contain only English examples but we test on all the languages. The results of both the experiments are reported in Table \ref{tab:mldoc}.

\begin{table}[h]
\centering
\resizebox{\linewidth}{!}{
\begin{tabular}{c|cc|cc|c}
\toprule
    & \multicolumn{2}{c|}{In language} & \multicolumn{3}{c}{Zero-shot} \\
\midrule
    & mBERT*          & MMTE          & mBERT*  & MMTE         & SOTA$^\dagger$ \\
\midrule
en  & 94.2           & \textbf{94.7}  & 94.2   & \textbf{94.7} & 89.9 \\
de  & 93.3           & \textbf{93.4}  & \textbf{80.2}   & 77.4          & 84.8 \\
zh  & 89.3           & \textbf{90.0}  & \textbf{76.9}   & 73.4          & 71.9 \\
es  & \textbf{95.7}  & 95.6           & 72.6   & \textbf{73.0} & 77.3 \\
fr  & \textbf{93.4}  & 92.7           & 72.6   & \textbf{77.2} & 78.0 \\
it  & \textbf{88.0}  & 87.6           & \textbf{68.9}   & 64.2          & 69.4 \\
ja  & \textbf{88.4}  & 88.1           & 56.5   & \textbf{69.0} & 60.3 \\
ru  & \textbf{87.5}  & 87.4           & \textbf{73.7}   & 68.9          & 67.8 \\
\midrule
avg & \textbf{91.2}  & 91.2           & 74.5   & \textbf{74.7} & 74.9 \\
\bottomrule
\end{tabular}}
\caption{Accuracies on the test set of the MLDoc dataset.  *mBERT numbers have been taken from \citet{wu2019beto}. mBERT is also the state-of-the-art for in-language training. $\dagger$ For zero-shot SOTA refers to \citet{artetxe2018massively}. \label{tab:mldoc}}
\end{table}

MMTE performance is on par with mBERT for in-language training on all the languages. It slightly edges over mBERT on zero-shot transfer while lagging behind SOTA by 0.2 points. Interestingly, MMTE beats SOTA on Japanese by more than 8 points. This may be due to the different nature and amount of data used for pre-training by these methods.

\subsection{Cross-lingual Intent Classification}
\citet{schuster2018cross} recently presented a dataset for multilingual task oriented dialog. This dataset contains 57k annotated utterances in English (43k), Spanish (8.6k), and Thai (5k) with 12 different intents across the domains weather, alarm, and reminder. The evaluation metric used is classification accuracy. We use this data for both in-language training and zero-shot transfer.  The task-specific network and the optimizer used is the same as the one used for the above two tasks. The learning rate schedule is (0.1,100k). Results are reported in Table \ref{tab:clic}. MMTE outperforms both mBERT and previous SOTA in both in-language and zero-shot setting on all 3 languages and establishes a new SOTA for this dataset.

\begin{table}[h]
\centering
\resizebox{\linewidth}{!}{
\begin{tabular}{c|ccc|ccc}
\toprule
    & \multicolumn{3}{c|}{In language} & \multicolumn{3}{c}{Zero-shot} \\
\midrule
    & mBERT*     & MMTE     & SOTA$^\dagger$    & mBERT*    & MMTE    & SOTA$^\dagger$    \\
\midrule
en  & 99.3      & \textbf{99.4}      & 99.1    & 99.3     & \textbf{99.4}     & 99.1    \\
es  & 98.4       & \textbf{98.8}      & 98.6    & 69.2      & \textbf{93.6}     & 85.4    \\
th  & 97.0       & \textbf{97.6}      & 97.4    & 43.4      & \textbf{89.6}     & 95.9    \\
\midrule
avg & 98.2       & \textbf{98.6}      & 98.4    & 70.6      & \textbf{94.2}     & 93.5    \\
\bottomrule
\end{tabular}}
\caption{Accuracies on the test set of the intent classification dataset. *mBERT numbers are from our own implementation using the publicly available mBERT checkpoint. $\dagger$ SOTA refers to the numbers reported in \citet{schuster2018cross}.} \label{tab:clic}
\end{table}
\subsection{POS Tagging}
We use universal dependencies POS tagging data from the Universal Dependency v2.3 \citep{nivre2018universal,zeman2018conll}. Gold segmentation is used for training, tuning and testing. The POS tagging task has 17 labels for all languages. We consider 48 different languages. These languages are chosen based on intersection of languages for which POS labels are available in the universal dependencies dataset and the languages supported by our mNMT model. The task-specific network consists of a one layer feed-forward neural network with 784 units. Since MMTE operates on the subword-level, we only consider the representation of the first subword token of each word. The optimizer used is Adafactor with learning rate schedule (0.1,40k). The evaluation metric used is F1-score, which is same as accuracy in our case since we use gold-segmented data. Results of both in-language and zero-shot setting are reported in Table \ref{tab:posr}.

While mBERT outperforms MMTE on in-language training by a small margin of 0.16 points, MMTE beats mBERT by nearly 0.6 points in the zero-shot setting. Similar to results in XNLI, we see MMTE outperform mBERT on low resource languages. Since mBERT is SOTA for zero-shot cross-lingual transfer on POS tagging task \cite{wu2019beto}, we also establish state-of-the-art on this dataset by beating mBERT in this setting.

\begin{table}[h]
\centering
\resizebox{.9\linewidth}{!}{
\begin{tabular}{c|cc|cc}
\toprule
    & \multicolumn{2}{c|}{In language} & \multicolumn{2}{c}{Zero-shot} \\
\midrule
    & mBERT*          & MMTE          & mBERT*         & MMTE         \\
\midrule
ar  & \textbf{97.25} & 96.72          & 58.67          & \textbf{66.87} \\
bg  & 98.86          & \textbf{98.93} & 86.65          & \textbf{87.61} \\
de  & \textbf{96.21} & 95.92          & \textbf{91.63} & 89.22          \\
en  & 96.11          & \textbf{96.34} & 96.16          & \textbf{96.34} \\
es  & 97.70          & \textbf{97.79} & \textbf{87.56} & 85.11          \\
fr  & \textbf{97.87} & 97.83          & \textbf{89.05} & 86.22          \\
hi  & 97.39          & \textbf{97.53} & 72.00          & \textbf{74.48} \\
mr  & 84.15          & \textbf{85.10} & 60.23          & \textbf{68.40} \\
nl  & \textbf{97.82} & 97.27          & \textbf{90.30} & 88.40          \\
pt  & \textbf{98.04} & 97.74          & \textbf{90.33} & 88.60          \\
ro  & \textbf{97.52} & 97.43          & \textbf{79.44} & 78.80          \\
sv  & \textbf{97.35} & 96.92          & \textbf{91.60} & 89.95          \\
te  & 93.66          & \textbf{93.73} & 81.12          & \textbf{83.40} \\
ur  & \textbf{94.43} & 94.42          & 65.88          & \textbf{69.36} \\
zh  & \textbf{96.25} & 95.87          & 67.38          & \textbf{67.51} \\
\midrule
48L & \textbf{95.65} & 95.49          & 74.87          & \textbf{75.43} \\
\bottomrule
\end{tabular}}
\caption{F1 scores on the dev set of the POS tagging dataset. *mBERT numbers are from our own implementation using the publicly available mBERT checkpoint. For readability, we only report results on 15 select languages and the 48 Language average (48L).} \label{tab:posr}
\end{table}
\subsection{Named Entity Recognition}
For NER, we use the dataset from the CoNLL 2002 and 2003 NER shared tasks, which when combined have 4 languages \cite{DBLP:journals/corr/cs-CL-0209010,sang2003introduction}. The labeling scheme is IOB with 4 types of named entities. The task-specific network, optimizer, and the learning rate schedule is the same as in the setup for POS tagging. The evaluation metric is span-based F1. Table \ref{tab:nerr} reports the results of both in-language and zero-shot settings.

MMTE performs significantly worse than mBERT on the NER task in all languages. On average, mBERT beats MMTE by 7 F1 points in the in-language setting and by more than 18 points in the zero-shot setting. We hypothesize that this might be because of two reasons: (1) mBERT is trained on clean Wikipedia data which is entity-rich while MMTE is trained on noisy web data with fewer entities, and (2) the translation task just copies the entities from the source to the target and therefore might not be able to accurately recognize them. This result points to the importance of the type of pre-training data and objective on down-stream task performance. We plan to investigate this further in future work.

\begin{table}[h]
\centering
\resizebox{.9\linewidth}{!}{
\begin{tabular}{c|cc|cc}
\toprule
    & \multicolumn{2}{c|}{In language} & \multicolumn{2}{c}{Zero-shot} \\
\midrule
    & mBERT*              & MMTE      & mBERT*             & MMTE     \\
\midrule
en  & \textbf{92.0}      & 87.8       & \textbf{92.0}     & 87.8      \\
es  & \textbf{87.4}      & 78.6       & \textbf{75.0}     & 52.0      \\
de  & \textbf{82.8}      & 78.7       & \textbf{69.6}     & 51.8      \\
nl  & \textbf{90.9}      & 81.7       & \textbf{77.6}     & 50.8      \\
\midrule
avg & \textbf{88.3}      & 81.7       & \textbf{78.5}     & 60.6     \\
\bottomrule
\end{tabular}}
\caption{F1 scores on the test set of the NER task. *mBERT numbers have been taken from \citet{wu2019beto}. mBERT is also the state-of-the-art for zero-shot setting.} \label{tab:nerr}
\end{table}

\section{Analysis} \label{sec: analysis}
In this section, we consider some additional settings for comparing mBERT and MMTE. We also investigate the impact of the number of languages and the target language token on MMTE performance.

\paragraph{Feature-based Approach}  
In this setting, instead of fine-tuning the entire network of mBERT or MMTE, we only fine-tune the task-specific network which only has a small percentage of the total number of parameters. The rest of the model parameters are frozen. We perform this experiment on POS tagging task by fine-tuning a single layer feed-forward neural network stacked on top of mBERT and MMTE. We report the results in Table \ref{tab:febert}. While the scores of the feature-based approach are significantly lower than those obtained via full fine-tuning (\ref{tab:posr}), we see that MMTE still outperforms mBERT on both in-language and zero-shot settings by an even bigger margin. This is particularly interesting as the feature-based approach has its own advantages: 1) it is applicable to downstream tasks which require significant task-specific parameters on top of a transformer encoder, 2) it is computationally cheaper to train and tune the downstream model, and 3) it is compact and scalable since we only need a small number of task-specific parameters.

\begin{table}[h]
\centering
\resizebox{.9\linewidth}{!}{
\begin{tabular}{c|cc|cc}
\toprule
    & \multicolumn{2}{c|}{In language} & \multicolumn{2}{c}{Zero-shot} \\
\midrule
    & mBERT          & MMTE          & mBERT         & MMTE         \\
\midrule
ar   & 87.65          & \textbf{91.50} & \textbf{61.08} & 60.35          \\
bg   & 95.82          & \textbf{96.72} & \textbf{88.08} & 83.48          \\
de   & 90.68          & \textbf{93.34} & 85.10          & \textbf{89.52} \\
en   & \textbf{92.73} & 92.15          & \textbf{92.73} & 92.33          \\
es   & \textbf{93.03} & 92.45          & 76.95          & \textbf{83.66} \\
fr   & 92.20          & \textbf{93.41} & 77.64          & \textbf{83.58} \\
hi   & 89.86          & \textbf{92.15} & 64.21          & \textbf{73.29} \\
mr   & 83.50          & \textbf{84.49} & 52.27          & \textbf{66.36} \\
nl   & 92.87          & \textbf{94.00} & 86.38          & \textbf{87.32} \\
pt   & 93.14          & \textbf{93.84} & 85.95          & \textbf{86.86} \\
ro   & \textbf{90.76} & 89.68          & 71.92          & \textbf{77.77} \\
sv   & 93.67          & \textbf{94.35} & 85.98          & \textbf{87.82} \\
te   & 90.57          & \textbf{92.75} & 70.09          & \textbf{81.12} \\
ur   & 90.02          & \textbf{90.32} & 58.98          & \textbf{67.90} \\
zh   & \textbf{86.29} & 84.34          & \textbf{60.82} & 56.50          \\
\midrule
avg  & 90.85          & \textbf{91.70} & 74.55          & \textbf{78.52} \\
\bottomrule
\end{tabular}}
\caption{F1 scores on dev set of POS tagging dataset using both mBERT and MMTE as feature extractor.} \label{tab:febert}
\end{table}

\paragraph{Few Shot Transfer}
While zero-shot transfer is a good measure of a model's natural cross-lingual effectiveness, the more practical setting is the few-shot transfer scenario as we almost always have access to, or can cheaply acquire, a small amount of data in the target language. We report the few-shot transfer results of mBERT and MMTE on the POS tagging dataset in \ref{tab:fews}. To simulate the few-shot setting, in addition to using English data, we use 10 examples from each language (upsampled to 1000). MMTE outperforms mBERT in few-shot setting by 0.6 points averaged over 48 languages. Once again, we see that the gains are more pronounced in low resource languages.

\begin{table}[h]
\centering
\resizebox{\linewidth}{!}{
\begin{tabular}{c|cc|c|cc}
\toprule
   & mBERT          & MMTE           &     & mBERT          & MMTE           \\
\midrule
ar & 85.20          & \textbf{85.53} & nl  & 90.01          & \textbf{90.26} \\
bg & \textbf{93.27} & 92.13          & pt  & 92.18          & \textbf{92.29} \\
de & \textbf{92.24} & 91.57          & ro  & 84.58          & \textbf{84.68} \\
en & 96.07          & \textbf{96.29} & sv  & 91.61          & \textbf{92.11} \\
es & \textbf{92.52} & 92.12          & te  & 83.84          & \textbf{84.74} \\
fr & \textbf{92.54} & 91.12          & ur  & 81.38          & \textbf{82.94} \\
hi & 86.29          & \textbf{87.32} & zh  & \textbf{81.14} & 76.32          \\
mr & 66.59          & \textbf{79.32} & 48L & 83.97          & \textbf{84.58} \\
\bottomrule
\end{tabular}}
\caption{F1 scores on dev set of POS tagging dataset in few-shot setting using 10 examples from each language in addition to English data.} \label{tab:fews}
\end{table}

\paragraph{One Model for all Languages}
Another setting of importance is the in-language training where instead of training one model for each language, we concatenate all the data and train one model jointly on all languages. We perform this experiment on the POS tagging dataset with 48 languages and report results in Table \ref{tab:onemodel}. We observe that MMTE performance is on par with mBERT. We also find that the 48 language average improves by 0.2 points as compared to the one model per language setting in Table \ref{tab:posr}.

\begin{table}[h]
\centering
\resizebox{\linewidth}{!}{
\begin{tabular}{c|cc|c|cc}
\toprule
   & mBERT          & MMTE           &     & mBERT          & MMTE           \\
\midrule
ar & \textbf{97.09} & 96.92          & nl  & \textbf{97.60} & 97.59          \\
bg & \textbf{98.83} & 98.94          & pt  & \textbf{98.42} & 98.02          \\
de & \textbf{95.94} & 95.68          & ro  & \textbf{97.48} & 97.47          \\
en & 96.04          & \textbf{96.28} & sv  & \textbf{97.27} & 97.22          \\
es & 97.71          & \textbf{97.77} & te  & \textbf{94.86} & 94.81          \\
fr & 97.74          & \textbf{97.86} & ur  & 94.39          & \textbf{94.77} \\
hi & 97.34          & \textbf{97.62} & zh  & \textbf{96.01} & 95.84          \\
mr & 84.70          & \textbf{86.13} & 48L & 95.67          & \textbf{95.69}  \\
\bottomrule
\end{tabular}}
\caption{F1 scores on dev set of POS tagging dataset in one model for all language setting.} \label{tab:onemodel}
\end{table}

\paragraph{Number of Languages in mNMT} We perform an ablation where we vary the number of languages used in the pre-training step. Apart from the 103 language setting, we consider 2 additional settings: 1) where we train mNMT on 4 languages to and from English, and 2) where we use 25 languages. The results are presented in Table \ref{tab:numlang}. We see that as we scale up the languages the zero-shot performance goes down on both POS tagging and XNLI tasks. These losses align with the relative BLEU scores of these models suggesting that the regressions are due to interference arising from the large number of languages attenuating the capacity of the NMT model. Scaling up the mNMT model to include more languages without diminishing cross-lingual effectiveness is a direction for future work.

\begin{table}[h]
\centering
\resizebox{\linewidth}{!}{
\begin{tabular}{c|ccc|ccc}
\toprule
   & \multicolumn{3}{c|}{POS} & \multicolumn{3}{c}{XNLI} \\
\midrule
   & 4x4   & 25x25 & 102x102 & 4x4   & 25x25  & 102x102 \\
\midrule
en & 96.30 & 96.25 & 96.28   & 80.1  & 80.0   & 79.6    \\
es & 86.61 & ----  & 85.13   & 74.2  & ----   & 71.6    \\
fr & 86.62 & 85.81 & 85.23   & 72.9  & 72.1   & 69.5    \\
de & 90.52 & 89.84 & 88.57   & 71.4  & 71.1   & 68.2    \\
zh & ----  & 67.83 & 67.51   & ----  & 72.7   & 69.2    \\
\bottomrule
\end{tabular}}
\caption{Effect of number of languages used for mNMT training on downstream zero-shot performance. XNLI numbers are accuracies. POS numbers are F1 scores. ---- indicate that the language was not present in the subset that mNMT was trained on.} \label{tab:numlang}
\end{table} 

\paragraph{Effect of the Target Language Token}
During the pre-training step, when we perform the translation task using the mNMT system, we prepend a $<$2xx$>$ token to the source sentence, where xx indicates the target language. The encoder therefore has always seen a $<$2en$>$ token in front of non-English sentences and variety of different tokens depending on the target language in front of English sentence. However, when fine-tuning on downstream tasks, we do not use this token. We believe this creates a mismatch between the pre-training and fine-tuning steps. To investigate this further, we perform a small scale study where we train an mNMT model on 4 languages to and from English in two different settings: 1) where we prepend the $<$2xx$>$ token, and  2) where we don't prepend the $<$2xx$>$ token but instead encode it separately. The decoder jointly attends over both the source sentence encoder and the $<$2xx$>$ token encoding. The BLEU scores on the translation tasks are comparable using both these approaches. The results on cross-lingual zero-shot transfer in both settings are provided in Table \ref{tab:2xxs}. Removing the $<$2xx$>$ token from the source sentence during mNMT training improves cross-lingual effectiveness on both POS tagging and XNLI task. Training a massively multilingual NMT model that supports translation of 102 languages to and from English without using the $<$2xx$>$ token in the encoder is another direction for future work.

\begin{table}[h]
\centering
\resizebox{\linewidth}{!}{
\begin{tabular}{c|cc|cc}
\toprule
    & \multicolumn{2}{c|}{POS}                           & \multicolumn{2}{c}{XNLI}                          \\
\midrule
    & with \textless{}2xx\textgreater{} & without       & with \textless{}2xx\textgreater{} & without       \\
\midrule
en  & 96.3                              & \textbf{96.4} & 80.1                              & \textbf{80.5} \\
es  & 85.1                              & \textbf{86.6} & 74.2                              & \textbf{74.5} \\
fr  & 86.6                              & \textbf{87.8} & 72.9                              & \textbf{73.7} \\
de  & 90.5                              & \textbf{91.1} & 71.4                              & \textbf{72.9} \\
\midrule
avg & 89.6                              & \textbf{90.5} & 74.6                              & \textbf{75.4} \\
\midrule
\end{tabular}}
\caption{Effect of $<$2xx$>$ token on zero-shot cross-lingual performance. XNLI numbers are accuracies. POS numbers are F1 scores.} \label{tab:2xxs}
\end{table}

\section{Related Work} \label{sec: rel_work}
We briefly review widely used approaches in cross-lingual transfer learning and some of the recent work in learning contextual word representations (CWR).

\paragraph{Multilingual Word Embeddings} For cross-lingual transfer, the most widely studied approach is to use multilingual word embeddings as features in neural network models. Several recent efforts have explored methods that align vector spaces for words in different languages \cite{faruqui2014improving,upadhyay2016cross,ruder2017survey}. 

\paragraph{Unsupervised CWR} More recent work has shown that CWRs obtained using unsupervised generative pre-training techniques such as language modeling or cloze task~\cite{taylor1953cloze} have led to state-of-the-art results beyond what was achieved with traditional word type representations on many monolingual NLP tasks \cite{peters2018deep,devlin2018bert,howard2018universal,radford2018improving} such as sentence classification, sequence tagging, and question answering. Subsequently, these contextual methods have been extended to produce multilingual representations by training a single model on text from multiple languages which have proven to be very effective for cross lingual transfer \cite{wu2019beto,mulcaire2019polyglot,piresmultilingual}. \citet{lample2019cross} show that adding a translation language modeling (TLM) objective to mBERT's MLM objective utilizes both monolingual and parallel data to further improve the cross-lingual effectiveness.

\paragraph{Representations from NMT} The encoder from an NMT model has been used as yet another effective way to contextualize word vectors \cite{mccann2017learned}. Additionally, recent progress in NMT has enabled one to train multilingual NMT systems that support translation from multiple source languages into multiple target languages within a single model~\cite{johnson2017google}. Our work is more closely related to two very recent works which explore the encoder from multilingual NMT model for cross-lingual transfer learning \cite{eriguchi2018zero,artetxe2018massively}. While \citet{eriguchi2018zero} also consider multilingual systems, they do so on a much smaller scale, training it on only 2 languages. \citet{artetxe2018massively} uses a large scale model comparable to ours with 93 languages but they constrain the model by pooling encoder representations and therefore only obtain a single vector per sequence. Neither of these approaches have been used on token level sequence tagging tasks. Further, neither concern themselves with the performance of the actual translation task whereas we our mNMT model performs comparable to bilingual baselines in terms of translation quality.

\section{Conclusion and Future Work}
\label{conclusion}
We train a massively multilingual NMT system using parallel data from 103 languages and exploit representations extracted from the encoder for cross-lingual transfer on various classification and sequence tagging tasks spanning over 50 languages. We find that the positive language transfer visible in improved translation quality for low resource languages is also reflected in the cross-lingual transferability of the extracted representations. The gains observed on various tasks over mBERT suggest that the translation objective is competitive with specialized approaches to learn cross-lingual embeddings.

We find that there is a trade off between the number of languages in the multilingual model and efficiency of the learned representations due to the limited capacity. Scaling up the model to include more languages without diminishing transfer learning capability is a direction for future work. Finally, one could also consider integrating mBERT's objective with the translation objective to pre-train the mNMT system.



\bibliographystyle{acl_natbib}
\bibliography{emnlp-ijcnlp-2019.bib}

\appendix

\section{Supplementary Material} \label{sec:supplemental}
In this section we provide the list of languages codes used throughout this paper and the statistics of the datasets used for the downstream tasks.

\begin{table*}[h]
\centering
\begin{tabular}{ll|ll|ll|ll}
\hline Language         & Id  & Language      & Id  & Language      & Id  & Language      & Id \\ \hline \hline
Afrikaans        & af & Galician         & gl & Latvian       & lv & Sindhi        & sd \\
Albanian         & sq & Georgian         & ka & Lithuanian    & lt & Sinhalese     & si \\
Amharic          & am & German           & de & Luxembouish   & lb & Slovak        & sk \\
Arabic           & ar & Greek            & el & Macedonian    & mk & Slovenian     & sl \\
Armenian         & hy & Gujarati         & gu & Malagasy      & mg & Somali        & so \\
Azerbaijani      & az & Haitian Creole    & ht & Malay         & ms & Spanish       & es \\
Basque           & eu & Hausa            & ha & Malayalam     & ml & Sundanese     & su \\
Belarusian       & be & Hawaiian        & haw & Maltese       & mt & Swahili       & sw \\
Bengali          & bn & Hebrew           & iw & Maori         & mi & Swedish       & sv \\
Bosnian          & bs & Hindi            & hi & Marathi       & mr & Tajik         & tg \\
Bulgarian        & bg & Hmong           & hmn & Mongolian     & mn & Tamil         & ta \\
Burmese          & my & Hungarian        & hu & Nepali        & ne & Telugu        & te \\
Catalan          & ca & Icelandic        & is & Norwegian     & no & Thai          & th \\
Cebuano         & ceb & Igbo             & ig & Nyanja        & ny & Turkish       & tr \\
Chinese          & zh & Indonesian       & id & Pashto        & ps & Ukrainian     & uk \\
Corsican         & co & Irish            & ga & Persian       & fa & Urdu          & ur \\
Croatian         & hr & Italian          & it & Polish        & pl & Uzbek         & uz \\
Czech            & cs & Japanese         & ja & Portuguese    & pt & Vietnamese    & vi \\
Danish           & da & Javanese         & jw & Punjabi       & pa & Welsh         & cy \\
Dutch            & nl & Kannada          & kn & Romanian      & ro & Xhosa         & xh \\
Esperanto        & eo & Kazakh           & kk & Russian       & ru & Yiddish       & yi \\
Estonian         & et & Khmer            & km & Samoan        & sm & Yoruba        & yo \\
Filipino/Tagalog  & tl & Korean           & ko & Scots Gaelic & gd   & Zulu          & zu \\
Finnish          & fi & Kurdish          & ku & Serbian       & sr & & \\
French           & fr & Kyrgyz           & ky & Sesotho       & st & & \\
Frisian          & fy & Lao              & lo & Shona         & sn & & \\ \hline
\end{tabular}
\caption{List of BCP-47 language codes used throughout this paper.}.
\label{tab:langids}
\end{table*}

\begin{table*}[]
\centering
\begin{tabular}{c|ccc}
\toprule
    & \#Training Samples & \#Dev Samples & \#Test Samples \\
\midrule
af  & 1315               & 194           & 425            \\
ar  & 6075               & 909           & 2643           \\
be  & 260                & 65            & 68             \\
bg  & 8907               & 1115          & 1116           \\
ca  & 13123              & 1709          & 1846           \\
cs  & 102993             & 11311         & 12203          \\
cu  & 4123               & 1073          & 1141           \\
da  & 4383               & 564           & 565            \\
de  & 13814              & 799           & 977            \\
el  & 1662               & 403           & 456            \\
en  & 19976              & 3777          & 3922           \\
es  & 28492              & 3054          & 2147           \\
et  & 24384              & 3125          & 3214           \\
eu  & 5396               & 1798          & 1799           \\
fa  & 4798               & 599           & 600            \\
fi  & 27198              & 3239          & 3422           \\
fr  & 18637              & 2902          & 1708           \\
fro & 13909              & 1842          & 1927           \\
gl  & 2872               & 860           & 861            \\
got & 3387               & 985           & 1029           \\
grc & 26491              & 2156          & 2353           \\
he  & 5241               & 484           & 491            \\
hi  & 13304              & 1659          & 1684           \\
hr  & 6983               & 849           & 1057           \\
hu  & 910                & 441           & 449            \\
id  & 4477               & 559           & 557            \\
ja  & 7133               & 511           & 551            \\
ko  & 27410              & 3016          & 3276           \\
la  & 34049              & 3335          & 3361           \\
lt  & 153                & 55            & 55             \\
lv  & 7163               & 1304          & 1453           \\
mr  & 373                & 46            & 47             \\
mt  & 1123               & 433           & 518            \\
nl  & 18058              & 1394          & 1472           \\
pt  & 17992              & 1770          & 1681           \\
ro  & 16008              & 1804          & 1781           \\
ru  & 53544              & 7163          & 7092           \\
sk  & 8483               & 1060          & 1061           \\
sl  & 8556               & 734           & 788            \\
sr  & 2935               & 465           & 491            \\
sv  & 7041               & 1416          & 2133           \\
ta  & 400                & 80            & 120            \\
te  & 1051               & 131           & 146            \\
tr  & 3685               & 975           & 975            \\
ug  & 1656               & 900           & 900            \\
uk  & 5290               & 647           & 864            \\
ur  & 4043               & 552           & 535            \\
zh  & 3997               & 500           & 500            \\
\bottomrule
\end{tabular}
\caption{Statistics of data used for the POS tagging task.}

\end{table*}

\begin{table*}[h]
\centering
\begin{tabular}{c|ccc}
\toprule
   & \#Training Samples & \#Dev Samples & \#Test Samples \\
\midrule
de & 12152              & 2867          & 3005           \\
en & 14041              & 3250          & 3453           \\
es & 8322               & 1914          & 1516           \\
ne & 15806              & 2895          & 5195          \\
\bottomrule
\end{tabular}
\caption{Statistics of data used for the NER task.}
\end{table*}

\begin{table*}[]
\centering
\begin{tabular}{c|ccc}
\toprule
   & \#Training Samples & \#Dev Samples & \#Test Samples \\
en & 392702             & 2490          & 5010           \\
ar & 0                  & 2490          & 5010           \\
bg & 0                  & 2490          & 5010           \\
de & 0                  & 2490          & 5010           \\
el & 0                  & 2490          & 5010           \\
es & 0                  & 2490          & 5010           \\
fr & 0                  & 2490          & 5010           \\
hi & 0                  & 2490          & 5010           \\
ru & 0                  & 2490          & 5010           \\
sw & 0                  & 2490          & 5010           \\
th & 0                  & 2490          & 5010           \\
tr & 0                  & 2490          & 5010           \\
ur & 0                  & 2490          & 5010           \\
vi & 0                  & 2490          & 5010           \\
zh & 0                  & 2490          & 5010           \\
\bottomrule
\end{tabular}
\caption{Statistics of data used for the XNLI task.}
\end{table*}

\begin{table*}[]
\centering
\begin{tabular}{c|ccc}
\toprule
   & \#Training Samples & \#Dev Samples & \#Test Samples \\
\midrule
en & 1000               & 1000          & 4000           \\
de & 1000               & 1000          & 4000           \\
zh & 1000               & 1000          & 4000           \\
es & 1000               & 1000          & 4000           \\
fr & 1000               & 1000          & 4000           \\
it & 1000               & 1000          & 4000           \\
ja & 1000               & 1000          & 4000           \\
ru & 1000               & 1000          & 4000           \\
\bottomrule
\end{tabular}
\caption{Statistics of data used for the document classification task.}
\end{table*}

\begin{table*}[]
\centering
\begin{tabular}{c|ccc}
\toprule
   & \#Training Samples & \#Dev Samples & \#Test Samples \\
   \midrule
en & 30521              & 4181          & 8621           \\
es & 3617               & 1983          & 3043           \\
th & 2156               & 1235          & 1692           \\
\bottomrule
\end{tabular}
\caption{Statistics of data used for the intent classification task.}
\end{table*}

\end{document}